\documentclass[runningheads,orivec]{llncs}
\usepackage[T1]{fontenc}

\usepackage{dsfont}
\usepackage{amsmath}
\usepackage{amssymb}
\usepackage{amsfonts}
\usepackage{nicematrix}
\usepackage{soul}
\usepackage{tikz}
\usetikzlibrary{arrows.meta}
\usetikzlibrary {patterns,patterns.meta}
\usetikzlibrary{backgrounds}
\usetikzlibrary{calc}
\usepackage{lipsum}
\usepackage{graphicx}
\usepackage[skins,breakable]{tcolorbox}
\usepackage{multirow}
\usepackage{array}
\newcolumntype{P}[1]{>{\centering\arraybackslash}p{#1}}
\newcolumntype{M}[1]{>{\centering\arraybackslash}m{#1}}
\usepackage[colorlinks=true,urlcolor=blue,linkcolor=blue,citecolor=blue]{hyperref}
\usepackage{ulem}
\usepackage{xcolor}
\newcommand{\approachname}{\textsc{SpaPool}}

\begin{document}
\title{{\approachname}: Soft Partition Assignment Pooling for Graph Neural Networks}

%
%
\author{
Rodrigue Govan\inst{1}\orcidID{0000-0002-4087-7056} \and
Romane Scherrer\inst{1}\orcidID{0000-0002-3020-6427} \and
Philippe Fournier-Viger\inst{2}\orcidID{0000-0002-7680-9899} \and
Nazha Selmaoui-Folcher\inst{1}\orcidID{0000-0003-1667-3819}
}
\authorrunning{
R. Govan et al.
}
%
\institute{
Institute of Exact and Applied Sciences (EA7484), University of New Caledonia \\
BP R4, F-98851 Nouméa, New Caledonia \\
\email{rodrigue.govan@gmail.com}, \email{nazha.selmaoui@unc.nc} \and
Big Data Institute, College of Computer Science and Software Engineering \\
Shenzhen University, China \\
\email{philfv@szu.edu.cn}
}
\maketitle              
\begin{abstract}
In deep learning, graph neural networks (GNNs) are powerful tools for processing complex, massive, and unstructured data represented as large graphs.
These graphs often feature interdependent entities, making GNNs particularly well-suited for analyzing such data.
However, the large number of nodes can present significant computational challenges, making efficient processing difficult.
To address this, several pooling methods have been proposed to reduce the size of graphs efficiently.
These methods fall into two categories: dense and sparse.
Dense pooling methods group nodes into a fixed number of clusters, predetermined by the user.
While this ensures a smaller graph size, it may not always be efficient, particularly with heterogeneous datasets.
Sparse pooling methods, on the other hand, select an adaptive number of significant nodes based on an importance score, removing non-significant nodes.
This adaptive approach allows for a more dynamic reduction of the graph, focusing on the most relevant nodes.
However, it can sometimes overlook the local and global structure of the input graph.
We introduce {\approachname}, a novel pooling method that combines the strengths of both dense and sparse techniques.
{\approachname} groups nodes into an adaptive number of clusters, leveraging the benefits of both approaches.
It aims to maintain the structural integrity of the graph while reducing its size efficiently.
Our experimental results on several datasets demonstrate that {\approachname} achieves competitive performance compared to existing pooling techniques and excels particularly on small-scale graphs.
This makes {\approachname} a promising method for applications requiring efficient and effective graph processing.
\end{abstract}

\section{Introduction}

Over the past few years, the advent of graph neural networks (GNNs)~\cite{Wu2021} has seen significant growth.
In opposite to standard deep learning methods such as recurrent neural networks (RNNs) and convolutional neural networks (CNNs)~\cite{Hoffmann2017} which require data structured as a table or a grid (e.g. pixels of an image), GNNs offer a wider flexibility to process unstructured data.
This flexibility makes GNNs particularly well-suited to various fields, such as biology, chemistry, economy and epidemiology, where instances from a dataset are not always independent.
These instances are characterized by attributes that are usually correlated with neighboring attributes, forming an important topological space that must be considered within a network with interdependent entities.

Graphs constitute a suited tool for modeling complex data.
However, processing high-dimensional graphs presents major challenges in terms of computational efficiency and preserving relevant structural information.
To address these challenges, pooling methods such as \textsc{DiffPool}~\cite{Ying2018} and \textsc{TopKPool}~\cite{Gao2019,Gao2021} have been proposed, aiming to reduce the size of graphs while preserving most essential information.
These pooling methods extract a subset of representative nodes, allowing the reduction of graphs without ignoring their informative structure.

Despite their achievements, these traditional approaches present a number of limitations when applied to heterogeneous graphs, which are characterized by strong variability in terms of size and topology.
Most existing pooling methods rely on two distinct approaches: they either group graph nodes into a fixed number of clusters (also called supernodes) or select an adaptive fraction of nodes based on an importance score with the remaining nodes being deleted.
In the first approach, node clustering allows the preservation of structural information from the input graph.
However, this clustering is based on a number of supernodes specified by the user, which can result in an output graph larger than the input graph, particularly in datasets where the number of nodes per graph is highly heterogeneous.
In the second approach, the user must set a fraction $k \in (0,1]$ prior to the GNN training.
This fraction $k$ represents the proportion of graph nodes to preserve for the pooled graph.
Therefore, graph nodes are assessed according to an importance score, and $\lceil kN \rceil$ most important nodes are retained ($N$ being the number of nodes in the input graph).
The remaining nodes are thus deleted.
This approach enables the reduction of each graph to be adapted according to its number of nodes.
Nevertheless, by deleting a subset of non-significant nodes, some structural information from the input graph may be lost.

In order to overcome these limitations, this paper proposes {\approachname} (Soft Partition Assignment Pooling), a new pooling method that combines the advantages of existing pooling approaches.
{\approachname} relies on a node clustering with an adaptive number of nodes determined according to the specific characteristics of each graph.

The article is organized as follows.
In the next section, a state of the art on graph neural networks and pooling methods is detailed.
The \autoref{sec:spapool} describes in detail the proposed {\approachname} method. 
The \autoref{sec:methods} details the applied methodology in order to test {\approachname} method and to compare it with existing pooling methods.
The \autoref{sec:results} presents experimental results obtained from several graph datasets, notably including a comparison with existing methods, but also with an ablation study that determines the effect of each {\approachname} component.
Finally, in the last section, a conclusion is drawn and some perspectives are discussed.

\section{Related Work}

Let $G = (V, E, X)$ be an attributed graph such that:
\begin{itemize}
    \item $V = \left\{ v_1, \dots, v_N \right\}$ is the set of $N$ nodes;
    \item $E \subseteq \{ (v_i, v_j) \in V^2 \mid \forall\: i,j \in \{1, \dots, N\} \wedge i \neq j \}$ is the set of $\lvert E \rvert$ edges;
    \item $X = \left\{ x_i \in \mathbb{R}^{F} \mid \forall\: i \in \{1, \dots, N\} \right\}$ is the attribute matrix where $x_i$ represents the vector of $F$ attributes associated to node $v_i$.
\end{itemize}

The adjacency matrix $A \in \mathbb{R}^{N \times N}$ is defined such that $A_{i,j} \neq 0$ if and only if $(v_i, v_j) \in E,\: \forall\: i,j$.

\subsection{Graph convolution}

To our knowledge, the first formal graph convolutional network (GCN) was proposed at the end of the last decade~\cite{Kipf2017}.
This model aimed to apply deep learning directly to graph-structured data, rather than transforming these graphs into tabular data, which could lead to a loss of information about the relationships between entities.
Inspired by the first order Laplacian method, the layer-wise propagation rule of a GCN is defined by:
\begin{equation}
    H^{(\ell+1)} = \sigma\left( \tilde{D}^{-\frac{1}{2}}\:\tilde{A}\:\tilde{D}^{-\frac{1}{2}}\:H^{(\ell)}\:W^{(\ell)} \right) \label{eq:gcn}
\end{equation}

\noindent where $\tilde{A} = A + I_N$ is the adjacency matrix with self-loops added using the identity matrix $I_N$, $H^{(\ell)}$ is the node embedding matrix at layer $\ell$, and $W^{(\ell)}$ is a trainable weight matrix that applies a linear transformation on the node embedding matrix.
Defined in $\{0,1\}^{N \times N}$, the identity matrix $I_N$ is described as a diagonal matrix where all diagonal elements are 1 and all off-diagonal elements are 0.
The diagonal degree matrix $\tilde{D}$ is defined such that $\tilde{D}_{i,i} = \sum\limits_{j=1}^{N} \tilde{A}_{i,j}$.
The function $\sigma(\cdot)$ is an activation function such as $\text{ReLU}(\cdot) = \max(0, \cdot)$.
Initially (i.e., $\ell = 0$), we have $H^{(0)} = X$.
Moreover, at layer $\ell > 0$, we are not referring to the node attribute matrix with the matrix $X$ anymore, but rather to the node embedding matrix with the matrix $H$.
This distinction is important since the input graph is processed through different layers of a GNN, therefore we do not have attributes describing nodes but embeddings.
Over the past few years, numerous GNN models have appeared.
Among them, \textsc{GraphSAGE}~\cite{Hamilton2017} integrates a sampling mechanism in order to generate node embeddings by iteratively aggregating information from adjacent nodes.
\textsc{GraphSAGE} is defined by:
\begin{equation}
    h_{i}^{(\ell + 1)} = \sigma\left( W^{(\ell)} \:.\:\text{AGG}\left( \left\{ h_j^{(\ell)},\: \forall j \in \mathcal{N}(i) \right\} \right)\right) \label{eq:graphsage}
\end{equation}
\noindent where $h_i^{(\ell)}$ represents embeddings of node $v_i$ at layer $\ell$, $\mathcal{N}(i)$ is the set of adjacent nodes to node $v_i$, $W^{(\ell)}$ is a trainable weight matrix, and AGG is an aggregation function such as the average, the sum, and the maximum.
Instead of aggregating information from every adjacent nodes simultaneously, \textsc{GraphSAGE} aggregates a subset of nodes fixed by the user, reducing the computational complexity and optimizing the modularity for processing high-dimensional graphs.
Another GNN model is the ``graph attention network'' (GAT)~\cite{Velickovic2018} that integrates an attention mechanism assigning different weight to adjacent nodes based on a calculated importance score function.
The GAT is defined by:
\begin{equation}
   h_{i}^{(\ell + 1)} = \sigma\left( \sum\limits_{j\:\in\:\mathcal{N}(i)} \alpha_{i,j}^{(\ell)}\:W^{(\ell)}\:h_j^{(\ell)} \right) \label{eq:gat}
\end{equation}
\noindent where $\sigma$ is an activation function such as ReLU, and $\alpha_{i,j}^{(\ell)}$ is the attention factor computed using the attention mechanism defined by:
\begin{equation}
    \alpha_{i,j}^{(\ell)} = \frac{\exp\left( \sigma\left(w^\top \left[W^{(\ell)} h_i^{(\ell)} \:\Vert\: W^{(\ell)} h_j^{(\ell)}\right] \right)\right)}{\sum\limits_{j \in \mathcal{N}(i)} \exp\left( \sigma\left(w^\top\left[W^{(\ell)} h_i^{(\ell)} \:\Vert\: W^{(\ell)} h_j^{(\ell)}\right] \right)\right)}
\end{equation}
\noindent where $w$ is a trainable weight vector, $\Vert$ represents the concatenation operator, and $\sigma$ is an activation function.
In this case, the activation function is $\text{LeakyReLU}(\cdot) = \max(0, \cdot) + \beta \times \min(0, \cdot)$ with $\beta$ a parameter defined by the user that corresponds to the negative slope.
Contrary to the GCN that considers all adjacent nodes with the same importance, the GAT emphasizes on the most important adjacent nodes, which can improve performance in the case where some relationships are more important than others~\cite{Luo2024}.
More recently, a model named ``graph isomorphism network'' (GIN)~\cite{Xu2019} has been proposed, employing an injective aggregation function in order to capture a richer structural information.
The GIN model is defined by:
\begin{equation}
    h_i^{(\ell + 1)} = \text{MLP}\left( (1+\epsilon)\:h_i^{(\ell)} + \sum\limits_{j\:\in\:\mathcal{N}(i)} h_j^{(\ell)} \right) \label{eq:gin}
\end{equation}
\noindent where $\epsilon$ is a trainable parameter allowing to control the importance of embeddings from the node itself compared to adjacent node embeddings.
This model was designed to be as efficient as the Weisfeiler-Lehman test~\cite{Huang2021}, which enables the distinction of isomorphic graphs from those that are not isomorphic.

All these GNNs have been developed in order to process graph-structured data.
They are focused on three main tasks: the node prediction, which aims to predict the label of certain nodes within a unique graph; the edge prediction, which consists of determining the existence of an edge (or a relationship) between nodes from a unique graph, and the graph prediction, which consists of predicting the label of an entire graph based on its attributes.
While GNNs have been developed for numerous tasks such as the graph matching task~\cite{Li2019} to compute a similarity score between two graphs, this paper focuses on graph classification task.
However, in numerous applications, graph sizes can be very large, making their processing complex, if not impossible.
Therefore, it becomes essential to reduce their size while preserving the most relevant information during a GNN training.

\subsection{Graph pooling}

In graph neural networks, implementing a pooling operator is more complex than in CNNs or RNNs as the data are not structured as a grid or a table, but rather as a connected network, with irregular dependencies and heterogeneous relationships between entities.
Therefore, defining a local region (or a neighborhood) for the pooling constitutes a real challenge.
In the literature, there exist two pooling approaches: the hierarchical pooling which consists of reducing the input graph into a smaller graph, and the global pooling which reduces the input graph into a single node.
This paper focuses on hierarchical pooling methods for their ability to maintain the topological graph structure, and we particularly focus on trainable pooling methods as they were specifically designed for GNNs.

In graph neural networks, the pooling process involves three steps:
\begin{enumerate}
    \item The selection of significant nodes by defining an operator $S$.
    The operator can be a matrix or a vector.
    In the case of a matrix, it is usually a sparse assignment matrix since $S$ allows to determine supernodes (i.e., centroids of node groups).
    This selection step is crucial in the pooling process in order to optimally reduce the input graph;
    \item The reduction of the node embedding matrix $H$ by using the operator $S$ ;
    \item The connection step which involves the readjustment of the adjacency matrix $A$ in order to maintain consistency with the node embedding matrix $H$.
\end{enumerate}

Among hierarchical trainable methods, pooling approaches can be divided into dense and sparse techniques.
Dense methods aim to group subsets of nodes into a fixed number of supernodes whose cardinality is $O(N)$.
These supernodes represent an aggregation of both local and global information.
For instance, \textsc{DiffPool}~\cite{Ying2018} realizes a hierarchical clustering and adjust the node clustering into supernodes during the GNN training.
\textsc{DiffPool} defines its matrix $S$ such that:
\begin{equation}
    S_{\textsc{DiffPool}} = \text{Softmax}\left( \text{GCN}\left(A^{(\ell)},H^{(\ell)}\right) \right) \label{eq:s-diffpool}
\end{equation}
In addition to the operator $S$, \textsc{DiffPool} also includes an auxiliary loss function in order to consider $S$ during the GNN training.
Therefore, the auxiliary loss function in \textsc{DiffPool} is defined by:
\begin{align}
    \mathcal{L}_{\textsc{DiffPool}} & = \mathcal{L}_{\text{LP}} + \mathcal{L}_{\text{E}} \label{eq:diffpool-loss} \\
    \mathcal{L}_{\text{LP}} & = {\left\lVert A^{(\ell)},\: S S^{\top} \right\rVert}_F \nonumber \\
    \mathcal{L}_{\text{E}} & = \frac{1}{N} \sum\limits_{i=1}^{N} E\left(S_i\right) \nonumber
\end{align}
\noindent where ${\lVert\cdot\rVert}_F$ corresponds to the Frobenius norm, and $E(\cdot)$ is the entropy function.

Although \textsc{DiffPool} is efficient to capture global structure from graphs, its computational complexity may lead to an overfitting in the case where the number of supernodes is not well calibrated.

Inspired by the minimum cut, \textsc{MinCUT}~\cite{Bianchi2020} is another dense method that groups nodes into supernodes by minimizing the number of inter-cluster connections while maximizing the intra-cluster connections.
This method captures structural information while preventing the over-grouping of uncorrelated nodes that can be produced by \textsc{DiffPool}.
\textsc{MinCUT} defines its matrix $S$ such that:
\begin{equation}
    S_{\textsc{MinCUT}} = \text{MLP}\left( H^{(\ell)} \right) \label{eq:s-mincut}
\end{equation}
Similarly to \textsc{DiffPool}, \textsc{MinCUT} also considers an auxiliary loss function.
This latter is defined by:
\begin{align}
    \mathcal{L}_{\textsc{MinCUT}} & = \underbrace{- \frac{\mathrm{Tr}(S^\top \tilde{A} S)}{\mathrm{Tr}(S^\top \tilde{D} S)}}_{\mathcal{L}_{c}} + \underbrace{{\left\lVert \frac{S^\top S}{{\left\lVert S^\top S \right\rVert}_F} - \frac{I_C}{\sqrt{C}} \right\rVert}_F}_{\mathcal{L}_{o}} \label{eq:mincut-loss}
\end{align}
\noindent where $C$ is a user-set parameter representing the number of supernodes in the output graph.

Through this auxiliary loss function, \textsc{MinCUT} aims to group highly connected nodes while ensuring that supernodes are orthogonal and of similar size.

\textsc{DMoN}~\cite{Tsitsulin2023} is another dense pooling method based on the information distribution throughout the input graph and its modularity.
Despite its operator $S$ defined as in \textsc{DiffPool} (\autoref{eq:s-diffpool}), \textsc{DMoN} integrates an auxiliary loss function that differs from the one in \textsc{DiffPool}.
The auxiliary loss function in \textsc{DMoN} is therefore defined by:
\begin{align}
    \mathcal{L}_{\textsc{DMoN}} & = -\frac{1}{2 \times \lvert E \rvert}\: \mathrm{Tr}(S^\top B S) + \frac{\sqrt{C}}{N} {\left\lVert \sum\limits_i S_i^\top \right\rVert}_F - 1 \label{eq:dmon-loss}
\end{align}
\noindent where $B = A - \frac{dd^T}{2 \times \lvert E \rvert}$ with $d$ the node degree vector, and $C$ the number of supernodes following the pooling layer.
While the loss function in \textsc{DiffPool} aims to group nodes with similar attributes while regularizing the node clustering entropy, the auxiliary loss function in \textsc{DMoN} aims to maximize the modularity of graphs.
Therefore, \textsc{DMoN} groups nodes according to their proximity in terms of attribute distributions, which allows to efficiently capture structural relationships in very large graphs.
This pooling method stands out from other pooling methods for its robustness in the case of complex and heterogeneous graphs.

We note that for these three dense pooling methods, their auxiliary loss function is respectively added to the classification loss function. In this case, it is the cross-entropy function, defined by:
\begin{equation}
    \mathcal{L}_{\text{CE}} = -\sum\limits_{i=1}^{N} y_i \log\left(\hat{y_i}\right) \label{eq:cross-entropy}
\end{equation}
\noindent where $y_i$ is the observed class of graph $i$, and $\hat{y_i}$ is the predicted probability of class $y_i$ of graph $i$.

Contrary to dense pooling methods, sparse pooling methods directly select a subset of nodes from the input graph based on criteria such as an importance score, which produces supernodes whose cardinality is $O(1)$.
To do so, one has to set a ratio $k\:\in\:(0, 1]$ indicating the proportion of nodes to retain.
These approaches explicitly reduce the dimension of a graph by removing non significant nodes while preserving its local structure.
Among these methods, \textsc{TopKPool}~\cite{Gao2019,Gao2021} stands out: it computes dynamically an importance score during the GNN training, preserving only nodes that have the highest score.
\textsc{TopKPool} defines its operator $S$ which is a vector such that :
\begin{equation}
    S_{\textsc{TopKPool}} = \frac{H^{(\ell)}p}{{\lVert p \rVert}_2}
\end{equation}
\noindent where $p \in \mathbb{R}^F$ is a trainable weight vector.

Nodes and edges that are retained aggregate the information.
The advantage of this approach is to simplify the input graph while dynamically adapting the most relevant node selection.
However, \textsc{TopKPool} may omit certain crucial information from the global structure of the input graph.
\textsc{SAGPool}~\cite{Lee2019} is another sparse pooling method, which employs a self-attention mechanism to calculate an importance score for each node within the input graph.
Therefore, it defines its matrix $S$ such that:
\begin{equation}
    S_{\textsc{SAGPool}} = \text{GCN}\left(A^{(\ell)}, H^{(\ell)} \right) \label{eq:s-sagpool}
\end{equation}
Similarly to the GIN model~\cite{Xu2019}, the self-attention enables the model to focus on most relevant nodes given their context within the graph.
The \textsc{SAGPool} method has this ability to be adaptable to different GNN architectures since it can be used with various convolution layers to compute importance scores.
While \textsc{SAGPool} focuses on most relevant nodes, another sparse pooling method named \textsc{ASAPool}~\cite{Ranjan2020}, focuses on the graph structure to conduct the pooling process.
This enables the preservation of structural information from the input graph.
Moreover, \textsc{ASAPool} employs the GNN training to compute the importance score of each node.
This score is not only based on the node attributes, but also on their context within the graph.

Once importance scores are computed, the majority of sparse pooling methods proceed in the same manner to determine which nodes to retain~\cite{Liu2022}, which is by selecting $\lceil kN \rceil$ nodes with the highest computed scores in $S_{\textsc{SAGPool}}$ and by removing rows (and columns) from the adjacency matrix that are associated to nodes that are not selected.

\section{{\approachname}: A Dense but Adaptive Pooling Approach}\label{sec:spapool}

\begin{figure}[ht]
    \centering
    \includegraphics[width=\linewidth, keepaspectratio]{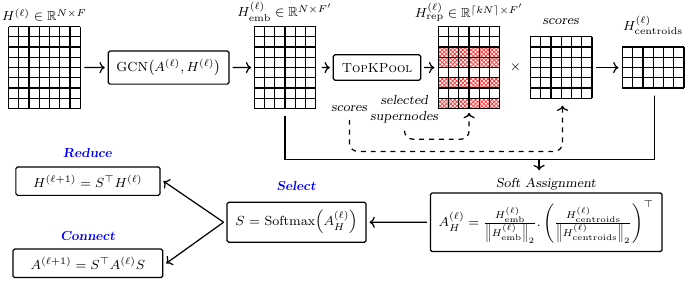}
    \caption{Illustration of the proposed graph pooling method ({\approachname}).}
    \label{fig:spapool}
\end{figure}

In this article, we propose {\approachname}~(Soft Partition Assignment Pooling), a graph pooling approach which combines node selection and association techniques to maximize the retained information during the graph reduction step (\autoref{fig:spapool}).
With {\approachname}, we aim to combine the benefits of dense and sparse methods simultaneously.
From dense methods, {\approachname} consists of grouping nodes into clusters (supernodes) in order to maintain local and global structures from the input graph contrary to existing sparse methods.
From sparse methods, {\approachname} consists of evaluating each node in order to select an adaptive ratio of nodes that will represent the supernodes.

\subsubsection*{Select.}

To define our operator $S$, we first proceeded with several steps as follows.

Given a layer $\ell$, the attributed graph can be represented by two matrices: the adjacency matrix $A^{(\ell)} \in \mathbb{R}^{N \times N}$ and the node embedding matrix $H^{(\ell)} \in \mathbb{R}^{N \times F}$.
Each non-zero cell in the adjacency matrix represents that two nodes in the graph are adjacent.
Each vector in row $h_i^{(\ell)}$ in the node embedding matrix denotes embeddings of node $v_i$ in the graph.
The layer-wise process of {\approachname} at layer $\ell$ proceeds as follows:

\begin{align}
    H_{\text{emb}}^{(\ell)} & = \text{GCN}\left(A^{(\ell)}, H^{(\ell)}\right) \nonumber \\
    H_{\text{rep}}^{(\ell)},\: H_{\text{scores}}^{(\ell)} & = \text{TopKPool}\left(H_{\text{emb}}^{(\ell)},\: k\right) && \text{with}\; k = 0.5 \nonumber \\
    H_{\text{centroids}}^{(\ell)} & = H_{\text{rep}}^{(\ell)}\: H_{\text{scores}} \nonumber \\
    S & = \text{Softmax}\left(\frac{H_{\text{emb}}^{(\ell)}}{{\left\lVert H_{\text{emb}}^{(\ell)} \right\rVert}_2}\:.\: \left(\frac{H_{\text{centroids}}^{(\ell)}}{{\left\lVert H_{\text{centroids}}^{(\ell)} \right\rVert}_2}\right)^{\top}\right) \label{eq:select-spapool}
\end{align}

At the end, the operator $S$ defined in {\approachname} (\autoref{eq:select-spapool}) is a Softmax function applied on the cosine similarity between the representative nodes and the node embeddings obtained from the GCN layer.

\subsubsection*{Reduce.}

In the reduction step, we readjust the attribute matrix as in existing pooling methods~\cite{Bianchi2020,Noutahi2019} which is defined by:
\begin{equation}
    H^{(\ell+1)} = S^\top H^{(\ell)}\label{eq:reduce}
\end{equation}

\subsubsection*{Connect.}

Finally, as in the reduction step, we set the connection step on adjacency matrix as in existing pooling methods~\cite{Bianchi2020,Noutahi2019,Ying2018} which is defined by:
\begin{equation}
    A^{(\ell+1)} = S^\top A^{(\ell)} S \label{eq;connect}
\end{equation}

\subsubsection*{Auxiliary loss.}

As {\approachname} is based on a dense approach (i.e., grouping nodes into supernodes), we considered an auxiliary loss function that we added into the classification loss function, i.e., the cross-entropy function (\autoref{eq:cross-entropy}).
Because we group nodes into supernodes, it can become difficult for our pooling method to avoid local minima issues.
Therefore, in the classification loss function, we added the same auxiliary losses used in \textsc{DiffPool}~\cite{Ying2018} that we recall as:
\begin{equation}
    \mathcal{L}_{\text{LP}} = {\left\lVert A^{(\ell)},\: S S^{\top} \right\rVert}_F \qquad\text{and}\qquad \mathcal{L}_{\text{E}} = \frac{1}{N} \sum\limits_{i=1}^{N} H\left(S_i\right) \nonumber
\end{equation}
\noindent where ${\lVert\cdot\rVert}_F$ denotes the Frobenius norm and $H(\cdot)$ is the entropy function.

\section{Methodology}\label{sec:methods}

In this methodology section, we are detailing the applied process in order to test {\approachname} and compare it with existing pooling methods.
This methodology notably includes the creation of a GNN model where the pooling layer is solely modified to conduct different tests.
Then, we will present datasets used to experiment {\approachname} and existing pooling methods. 

\subsection{Graph Neural Network}

\begin{figure}
    \centering
    \includegraphics[width=\linewidth, keepaspectratio]{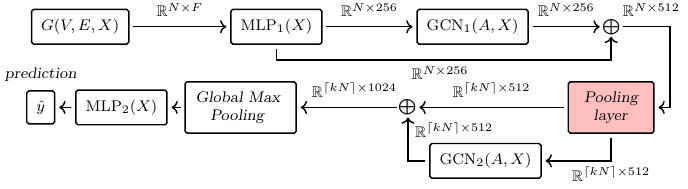}
    \caption{Graph Neural Network model used in the experiments.}
    \label{fig:gnn-model}
\end{figure}

In this section, we define the graph neural network to compare {\approachname} with other pooling approaches from the literature.
To do so, we designed a unique graph neural network model in order to evaluate the impact of the pooling layer solely.
The GNN is composed of two MLP blocks and two GCN blocks as illustrated in the \autoref{fig:gnn-model}.
Only one layer concern the pooling layer which allows us to switch {\approachname} with existing pooling methods.

\subsection{Datasets}

For each dataset (\autoref{tab:datasets}), it has been randomly divided into a training, validation and test sets representing 80\%, 10\%, and 10\% of the dataset, respectively.
All models were trained, evaluated, and tested over the same sets in order to obtain consistent and reproducible results.
The training hyper-parameters that were used are the following: a batch size of 64 graphs, a learning rate of $\eta = 5 \times 10^{-4}$, a pooling ratio of $k = 0.5$, an optimization via the stochastic gradient descent (SGD), and a maximum number of 500 epochs with an early stopping set at 100 epochs.
The loss function used for the classification task is the cross-entropy function (\autoref{eq:cross-entropy}).
All experiments were conducted using Python 3.8 with \texttt{PyTorch} 1.13.1 and \texttt{PyTorch-Geometric} 2.6.1 libraries.

\begin{table}[ht]
    \centering
    \caption{Summary of datasets~\cite{Morris2020} used in our graph classification experiments.}
    \label{tab:datasets}
    \begin{tabular}{|p{0.25\linewidth}|M{0.1\linewidth}|M{0.165\linewidth}|M{0.185\linewidth}|M{0.125\linewidth}|M{0.1\linewidth}|}
        \hline
        Dataset & Graphs & Nodes (avg) & Edges (avg) & Attributes & Classes \\
        \hline
        \texttt{PROTEINS} & 1,113 & $39.06_{\pm 45.76}$ & $72.82_{\pm 84.6}$ & 4 & 2 \\ 
        \texttt{ENZYMES} & 600 & $32.63_{\pm 15.28}$ & $62.14_{\pm 25.5}$ & 21 & 6 \\ 
        \texttt{DD} & 1,178 & $284.32_{\pm 272}$ & $715.66_{\pm 693.91}$ & 89 & 2 \\ 
        \texttt{Mutagenicity} & 4,337 & $30.32_{\pm 20.12}$ & $30.77_{\pm 16.82}$ & 14 & 2 \\ 
        \texttt{github\_stargazers} & 12,725 & $113.79_{\pm 164}$ & $234.64_{\pm 427.23}$ & 1 & 2 \\ 
        \texttt{reddit\_threads} & 203,088 & $23.93_{\pm 16.55}$ & $24.85_{\pm 19.14}$ & 1 & 2 \\
        \texttt{OHSU} & 79 & $82.01_{\pm 43.44}$ & $199.66_{\pm 165.08}$ & 190 & 2 \\
        \texttt{twitch\_egos} & 127,094 & $29.67_{\pm 11.1}$ & $86.59_{\pm 70.37}$ & 1 & 2 \\
        \texttt{COLLAB} & 5,000 & $74.49_{\pm 62.3}$ & $2457.22_{\pm 6438.92}$ & 1 & 3 \\
        \texttt{IMDB-BINARY} & 1,000 & $19.77_{\pm 10.06}$ & $96.53_{\pm 105.6}$ & 1 & 2 \\
        \hline
    \end{tabular}
\end{table}

These datasets were chosen for their variety in terms of number of graphs, nodes, attributes and classes.

\section{Results}\label{sec:results}

In this section, we present the results obtained from tested datasets (\autoref{tab:datasets}) as well as an ablation study of {\approachname} components.
An ablation study consists of assessing the contribution of each component within the proposed method in the prediction performance.
This type of study is commonly used, particularly for a GNN pooling in which the pooling method is characterized by a sequence of graph processing~\cite{Gao2019,Gao2021,Ranjan2020}.

\subsection{Performance Comparison}

To experiment {\approachname} on the datasets (\autoref{tab:datasets}) mentioned in the previous section, we trained a GNN model (\autoref{fig:gnn-model}) ten times, with ten different random split (train, validation, test) in order to obtain different accuracy results.
All experiments were conducted on a high-performance computing in which we limited each training on one NVIDIA Tesla V100 GPU with 32GB of dedicated memory.
The \autoref{tab:results} reports the average and standard deviation of the graph classification accuracy on the test set.

\begin{table}[ht]
    \centering
    \caption{Accuracies on the graph classification experiments.}
    \label{tab:results}
    \begin{tabular}{|p{0.25\linewidth}|M{0.15\linewidth}M{0.15\linewidth}M{0.15\linewidth}|M{0.15\linewidth}|}
        \hline
        Dataset & \textsc{TopKPool}   & \textsc{ASAPool}   & \textsc{SAGPool}   & \textsc{SpaPool} \\
        \hline
        \texttt{PROTEINS} & $68.21_{\pm 11.31}$ & $\mathbf{71.51_{\pm 8.88}}$ & $70.62_{\pm 9.52}$ & $71.24_{\pm 7.53}$ \\
        \texttt{ENZYMES} & $62.50_{\pm 4.84}$  & $65.62_{\pm 3.76}$ & $55.36_{\pm 5.93}$ & $\mathbf{68.29_{\pm 5.31}}$ \\
        \texttt{DD} & $74.57_{\pm 5.27}$  & $76.01_{\pm 3.88}$ & $\mathbf{77.03_{\pm 4.03}}$ & $72.46_{\pm 0.45}$ \\
        \texttt{Mutagenicity} & $75.04_{\pm 3.23}$  & $76.10_{\pm 1.83}$ & $75.36_{\pm 3.91}$ & $\mathbf{76.58_{\pm 1.91}}$ \\
        \texttt{github\_stargazers} & $65.32_{\pm 1.63}$  & $\mathbf{67.8_{\pm 1.28}}$  & $67.26_{\pm 2.07}$ & $64.83_{\pm 3.58}$ \\
        \texttt{reddit\_threads} & $75.82_{\pm 1.63}$  & $\mathbf{77.09_{\pm 0.53}}$ & $76.67_{\pm 1.03}$ & $76.09_{\pm 0.8}$ \\
        \texttt{OHSU} & $61.25_{\pm 17.18}$ & $55_{\pm 18.71}$   & $\mathbf{62.5_{\pm 14.79}}$ & $53.75_{\pm 20.19}$ \\
        \texttt{twitch\_egos} & $66.98_{\pm 2.75}$  & $69.68_{\pm 0.8}$  & $\mathbf{70.03_{\pm 0.41}}$ & $68.56_{\pm 1.45}$ \\
        \texttt{COLLAB} & $66.26_{\pm 3.5}$   & $67.07_{\pm 3.5}$  & $\mathbf{68.39_{\pm 2.92}}$ & $54.53_{\pm 2.13}$ \\        
        \texttt{IMDB-BINARY} & $64.1_{\pm 4.99}$   & $68.7_{\pm 3.29}$  & $\mathbf{69.1_{\pm 4.55}}$  & $55.9_{\pm 3.56}$ \\
        \hline
    \end{tabular}
\end{table}

Although our {\approachname} method outperformed existing methods in 2 out of the 10 datasets tested, we still achieved equivalent results with minimal variability across ten training sessions.
The only strong variability observed in all tested pooling methods was with \texttt{OHSU} dataset, which is the smallest dataset (\autoref{tab:datasets}) in our tests with only 79 graphs in the dataset.

If we look more closely at the results, as {\approachname} is based on the \textsc{TopKPool} method to score graph nodes, our pooling method outperformed in 5 out of the 10 tested datasets.
More specifically, {\approachname} obtained better results than \textsc{TopKPool} in datasets where graphs are smaller (i.e., around 30 nodes per graph in average).
This difference is notably due to the fact that \textsc{TopKPool} removed non-significant nodes during the pooling process, while {\approachname} grouped nodes into supernodes.
This grouping indeed preserved the information of the grouped nodes, while \textsc{TopKPool} removed that information.
In datasets where there is a very small number of nodes per graph, those nodes become essential for understanding the graph.
Therefore, using a sparse pooling method that removes nodes such as \textsc{TopKPool} may be less effective than using a dense method that groups nodes.

In the next subsections, we provide an ablation study of the different components of {\approachname}.
The ablation study consists of examining the contribution of each component within the proposed method to prediction performance.
In this case, we assessed the contribution of the node selection component (i.e., \textsc{TopKPool}), the node aggregation (i.e., the cosine similarity), and the loss function used.

\subsection{Effect of node selection}

Following the graph classification performance, we assessed the importance of the node selection method in {\approachname} by comparing \textsc{TopKPool} and \textsc{SAGPool}.
\textsc{SAGPool} employs a GCN layer (\autoref{eq:gcn}) on the node embedding matrix prior to the significant node identification while \textsc{TopKPool} multiplies the node embedding matrix by a normalized weight vector.

\begin{table}[ht]
    \centering
    \caption{Graph classification accuracies according to node selection methods.}
    \label{tab:select}
    \begin{tabular}{|p{0.25\linewidth}|M{0.25\linewidth}|M{0.25\linewidth}|}
        \hline
        Dataset               & \textsc{TopKPool}   & \textsc{SAGPool} \\
        \hline
        \texttt{PROTEINS}     & $\mathbf{71.24_{\pm 7.53}}$ & $68.21_{\pm 12.24}$ \\
        \texttt{ENZYMES}      & $\mathbf{68.29_{\pm 5.31}}$ & $66.25_{\pm 4.96}$ \\
        \texttt{Mutagenicity} & $76.58_{\pm 1.91}$ & $\mathbf{77.72_{\pm 1.91}}$ \\
        \texttt{OHSU}         & $\mathbf{53.75_{\pm 20.19}}$ & $51.25_{\pm 18.92}$ \\
        \texttt{IMDB-BINARY}  & $\mathbf{55.9_{\pm 3.56}}$ & $55.7_{\pm 3.61}$ \\
        \hline
    \end{tabular}
\end{table}

As shown in \autoref{tab:select}, the use of \textsc{TopKPool} appeared to optimize performance compared to \textsc{SAGPool}, as the latter decreased accuracies in the tested datasets.

\subsection{Effect of node aggregation}

In our ablation study, we further assessed the importance of the node aggregation method by comparing the retained method (cosine similarity), the scalar product, and the attention mechanism.

In our tests, the scalar product is defined as:
\begin{align}
    A_H^{(\ell)} & = \text{Softmax}\left( H^{(\ell)}\:H_{\text{rep}}^{(\ell)^\top} \right) \nonumber
\end{align}

At layer $\ell$, the attention mechanism is defined as:
\begin{align}
    H_q^{(\ell)} & = \text{MLP}\left(H^{(\ell)}\right) \nonumber \\
    H_k^{(\ell)} & = \text{MLP}\left( H_{\text{rep}}^{(\ell)} \right) \nonumber \\
    A_H^{(\ell)} & = \text{Softmax}\left( \frac{1}{\sqrt{\alpha}}\: H_q^{(\ell)}\:H_k^{(\ell)^\top} \right) \nonumber
\end{align}

\noindent where we set $\alpha = 16$, $H_q^{(\ell)} \in \mathbb{R}^{N \times \alpha}$, and $H_k^{(\ell)} \in \mathbb{R}^{N \times \alpha}$.

\begin{table}[ht]
    \centering
    \caption{Graph classification accuracies according to the node aggregation method.}
    \label{tab:asso}
    \begin{tabular}{|p{0.25\linewidth}|M{0.175\linewidth}|M{0.175\linewidth}|M{0.175\linewidth}|}
        \hline
        Dataset & \textsc{Cosine}   & \textsc{Scalar} & \textsc{Attention} \\
        \hline
        \texttt{PROTEINS} & $\mathbf{71.24_{\pm 7.53}}$ & $71.07_{\pm 9.78}$ & $66.16_{\pm 12.26}$ \\
        \texttt{ENZYMES} & $68.29_{\pm 5.31}$ & $\mathbf{69.64_{\pm 6.99}}$ & $69.09_{\pm 5.98}$ \\
        \texttt{Mutagenicity} & $76.58_{\pm 1.91}$ & $77.35_{\pm 1.89}$ & $\mathbf{77.6_{\pm 2.2}}$ \\
        \texttt{OHSU} & $\mathbf{53.75_{\pm 20.19}}$ & $47.50_{\pm 20.77}$ & $43.75_{\pm 21.1}$ \\
        \texttt{IMDB-BINARY} & $\mathbf{55.9_{\pm 3.56}}$ & $55.70_{\pm 4.15}$ & $54.10_{\pm 3,33}$ \\
        \hline
    \end{tabular}
\end{table}

In most of the experimented datasets (\autoref{tab:asso}), the cosine similarity outperformed both the scalar product and the attention mechanism.
Moreover, the cosine similarity resulted in lower variability compared to the two other aggregation methods.

\subsection{Effect of the auxiliary loss function}

Finally, in our ablation study, we assessed the auxiliary loss function.
While {\approachname} employs the auxiliary loss function from \textsc{DiffPool} (\autoref{eq:diffpool-loss}), we also tested auxiliary loss functions from \textsc{DMoN} (\autoref{eq:dmon-loss}) and \textsc{MinCUT} (\autoref{eq:mincut-loss}).
As \textsc{MinCUT} is a dense pooling method, the constant $C$ in $\mathcal{L}_{\textsc{MinCUT}}$ is the number of supernodes defined by the user to obtain in the pooling output graph.
As {\approachname} is an adaptive pooling method, we modified $\mathcal{L}_{o}$ from \autoref{eq:mincut-loss} as:
\begin{equation}
    \mathcal{L}_{o} = \frac{1}{N} \sum\limits_{i=1}^{N} \left\lVert \frac{S_i^\top\:S_i}{\lVert S_i^\top\:S_i\rVert} - \frac{I_{k_i}}{\sqrt{k_i}} \right\rVert
\end{equation}
\noindent where $k_i$ is the number of supernodes in the graph $i$.
The same modification was applied regarding the auxiliary loss function in \textsc{DMoN} (\autoref{eq:dmon-loss}).

\begin{table}[ht]
    \centering
    \caption{Graph classification accuracies according to the auxiliary loss function.}
    \label{tab:loss}
    \begin{tabular}{|p{0.25\linewidth}|M{0.175\linewidth}|M{0.175\linewidth}|M{0.175\linewidth}|}
        \hline
        Dataset & $\mathcal{L}_{\textsc{DiffPool}}$ & $\mathcal{L}_{\textsc{DMoN}}$ & $\mathcal{L}_{\textsc{MinCUT}}$ \\
        \hline
        \texttt{PROTEINS}     & $\mathbf{71.24_{\pm 7.53}}$ & $66.07_{\pm 12.32}$ & $68.75_{\pm 10.68}$ \\
        \texttt{ENZYMES}      & $68.29_{\pm 5.31}$ & $68.81_{\pm 5.93}$ & $\mathbf{70.82_{\pm 5.39}}$ \\
        \texttt{Mutagenicity} & $76.58_{\pm 1.91}$ & $\mathbf{78.34_{\pm 2.1}}$ & $76.8_{\pm 1.79}$ \\
        \texttt{OHSU}         & $53.75_{\pm 20.19}$ & $\mathbf{53.75_{\pm 17.72}}$ & $52.5_{\pm 18.37}$ \\
        \texttt{IMDB-BINARY}  & $55.9_{\pm 3.56}$ & $54.5_{\pm 3.38}$ & $\mathbf{55.9_{\pm 3.21}}$ \\
        \hline
    \end{tabular}
\end{table}

In our {\approachname} ablation study, the effect of the auxiliary loss function remains the least unanimous, as none of the tested auxiliary loss functions clearly outperformed the others in our datasets (\autoref{tab:loss}).
We experimented with {\approachname} on several datasets that differ in terms of the number of graphs and the number of attributes.
Determining a generic auxiliary loss function for any type of graph dataset remains a challenging task in graph neural networks.

\section{Conclusion}

While most pooling methods in GNNs are either dense and fixed or sparse and adaptive, this paper proposed a novel method named {\approachname}, a dense yet adaptive pooling approach that leverages the benefits of both dense and sparse methods.
To do so, we detailed the process and experimented our pooling method on 10 different datasets, and we compared it with existing methods.
Although {\approachname} did not outperform existing methods in all datasets, it still managed to obtain comparable results, which indicates a great potential for our method to optimize the process.
Additionally, our method obtained satisfying results in datasets where graphs are small (i.e., an average of 30 nodes per graph) as it grouped nodes into supernodes to preserve information, as well as the local and global graph structure.

Next steps of our work will include adjustment in order to obtain satisfying results in large graphs, but also in attributed graphs with a large number of attributes.
Future works also include an explainability and an interpretability components that will enable the understanding of which features from the attributes are considered as important according to the trained GNN.

\begin{credits}

\subsubsection{\ackname} This work was funded by the French National Research Agency as part of the SPIraL program (grant number ANR-19-CE35-0006-02).

\subsubsection{\discintname}
The authors declare that no competing interests exist.

\end{credits}
%
%
%
\bibliographystyle{splncs04}
\bibliography{bibliography}

\begin{thebibliography}{10}
\providecommand{\url}[1]{\texttt{#1}}
\providecommand{\urlprefix}{URL }
\providecommand{\doi}[1]{https://doi.org/#1}

\bibitem{Bianchi2020}
Bianchi, F.M., Grattarola, D., Alippi, C.: Spectral clustering with graph neural networks for graph pooling. In: Proceedings of the 37th International Conference on Machine Learning. pp. 874--883. PMLR (2020)

\bibitem{Gao2019}
Gao, H., Ji, S.: Graph u-nets. In: Proceedings of the 36th International Conference on Machine Learning. vol.~97, pp. 2083--2092. PMLR (2019)

\bibitem{Gao2021}
Gao, H., Ji, S.: Graph u-nets. IEEE Transactions on Pattern Analysis and Machine Intelligence pp. 4948--4960 (2021). \doi{10.1109/tpami.2021.3081010}

\bibitem{Hamilton2017}
Hamilton, W.L., Ying, R., Leskovec, J.: Inductive representation learning on large graphs. In: Proceedings of the 31st International Conference on Neural Information Processing Systems. pp. 1025--1035. NIPS (2017)

\bibitem{Hoffmann2017}
Hoffmann, J., Navarro, O., Kastner, F., Jan{\ss}en, B., Hubner, M.: A survey on cnn and rnn implementations. In: Proceedings of the 7th International Conference on Performance, Safety and Robustness in Complex Systems and Applications. PESARO (2017)

\bibitem{Huang2021}
Huang, N.T., Villar, S.: A short tutorial on the weisfeiler-lehman test and its variants. In: IEEE International Conference on Acoustics, Speech and Signal Processing. pp. 8533--8537. ICASSP (2021). \doi{10.1109/ICASSP39728.2021.9413523}

\bibitem{Kipf2017}
Kipf, T.N., Welling, M.: Semi-supervised classification with graph convolutional networks. In: International Conference on Learning Representations. ICLR (2017)

\bibitem{Lee2019}
Lee, J., Lee, I., Kang, J.: Self-attention graph pooling. In: Proceedings of the 36th International Conference on Machine Learning. pp. 3734--3743. PMLR (2019)

\bibitem{Li2019}
Li, Y., Gu, C., Dullien, T., Vinyals, O., Kohli, P.: Graph matching networks for learning the similarity of graph structured objects. In: Proceedings of the 36th International Conference on Machine Learning. pp. 3835--3845. PMLR (2019)

\bibitem{Liu2022}
Liu, C., Zhan, Y., Li, C., Du, B., Wu, J., Hu, W., Liu, T., Tao, D.: Graph pooling for graph neural networks: Progress, challenges, and opportunities. arXiv preprint arXiv:2204.07321  (2022). \doi{10.48550/arXiv.2204.07321}

\bibitem{Luo2024}
Luo, Y., Shi, L., Wu, X.M.: Classic {GNN}s are strong baselines: Reassessing {GNN}s for node classification. In: Proceedings of the 38th International Conference on Neural Information Processing Systems. NIPS (2024)

\bibitem{Morris2020}
Morris, C., Kriege, N.M., Bause, F., Kersting, K., Mutzel, P., Neumann, M.: Tudataset: A collection of benchmark datasets for learning with graphs. In: ICML 2020 Workshop on Graph Representation Learning and Beyond (GRL+ 2020) (2020), \url{www.graphlearning.io}

\bibitem{Noutahi2019}
Noutahi, E., Beaini, D., Horwood, J., Giguère, S., Tossou, P.: Towards interpretable sparse graph representation learning with laplacian pooling. arXiv preprint arXiv:1905.11577  (2019). \doi{10.48550/arXiv.1905.11577}

\bibitem{Ranjan2020}
Ranjan, E., Sanyal, S., Talukdar, P.: Asap: Adaptive structure aware pooling for learning hierarchical graph representations. In: Proceedings of the 34th AAAI Conference on Artificial Intelligence. pp. 5470--5477. AAAI (2020). \doi{10.1609/aaai.v34i04.5997}

\bibitem{Tsitsulin2023}
Tsitsulin, A., Palowitch, J., Perozzi, B., M{\"u}ller, E.: Graph clustering with graph neural networks. Journal of Machine Learning Research  \textbf{24}(127),  1--21 (2023)

\bibitem{Velickovic2018}
Veličković, P., Cucurull, G., Casanova, A., Romero, A., Liò, P., Bengio, Y.: Graph attention networks. In: International Conference on Learning Representations (2018)

\bibitem{Wu2021}
Wu, Z., Pan, S., Chen, F., Long, G., Zhang, C., Yu, P.S.: A comprehensive survey on graph neural networks. IEEE Transactions on Neural Networks and Learning Systems  \textbf{32}(1),  4--24 (2021). \doi{10.1109/tnnls.2020.2978386}

\bibitem{Xu2019}
Xu, K., Hu, W., Leskovec, J., Jegelka, S.: How powerful are graph neural networks? In: International Conference on Learning Representations. ICLR (2019)

\bibitem{Ying2018}
Ying, Z., You, J., Morris, C., Ren, X., Hamilton, W., Leskovec, J.: Hierarchical graph representation learning with differentiable pooling. In: Proceedings of the 32nd International Conference on Neural Information Processing Systems. NIPS (2018)

\end{thebibliography}

\end{document}